\documentclass[journal]{IEEEtran}
\usepackage[T1]{fontenc}
\usepackage{amsmath,amssymb}
\usepackage{graphicx}
\usepackage{booktabs}
\usepackage{cite}
\usepackage{xspace}
\usepackage{tikz}
\usetikzlibrary{arrows.meta,positioning,calc,fit,backgrounds}
\usepackage{pgfplots}
\usepackage{pgfplotstable}
\pgfplotsset{compat=1.17}
\usepgfplotslibrary{groupplots}
\usepackage{multirow} 
\usepackage{caption}
\usepackage{makecell}
\usepackage{adjustbox}
\usepackage{tikz}
\usepackage{url}
\usepackage[disable]{todonotes}
\usepackage{xcolor}
\newcommand{\rev}[1]{\textcolor{red}{#1}}
\usepackage{todonotes}
\usepackage{subcaption}

\usepackage{xcolor}
\usepackage{hyperref}

\hypersetup{
    colorlinks=true,
    citecolor=black,
    linkcolor=black,
    urlcolor=blue
}

\colorlet{physc}{teal!85!black}
\colorlet{cmdc}{orange!90!black}
\colorlet{actc}{violet!80!black}
\colorlet{colB}{orange!85!black}
\colorlet{colM}{blue!70!black}
\colorlet{colP}{green!50!black}
\colorlet{polc}{teal!75!blue}
\colorlet{exec}{orange!85!black}

\newcommand{\spindle}{\textsc{Spindle}}

\AtBeginDocument{
    \setlength{\abovedisplayskip}{2pt}
    \setlength{\belowdisplayskip}{2pt}
    \setlength{\abovedisplayshortskip}{2pt}
    \setlength{\belowdisplayshortskip}{2pt}
}

\begin{document}

\title{SPINDLE: Real-Time Proprioceptive Correction for Execution-Robust Action-Chunked VLAs}

% OR 

% SPINDLE: Closing the Execution Loop in Action-Chunked VLA Policies via Lightweight Proprioceptive Correction

\author{Farida Mohsen$^{*}$,~\IEEEmembership{Member,~IEEE,}
        Thowayba Elkaffash, Hasan Kurban, Mohamed Mabrok, Nader Meskin,~\IEEEmembership{Senior Member,~IEEE,}
        and~Ali Safa,~\IEEEmembership{Member,~IEEE}% <- this % stops a space
\thanks{F. Mohsen developed the presented methods, designed and implemented the experiments, contributed to the real-world experimental deployment, and contributed to the result analysis, and wrote the first draft of the paper; T. Elkaffash conceived the real-world experimental setup, contributed to the data collection, and performed the experiments using the real-world setup; H. Kurban, M. Mabrok, N. Meskin and A. Safa contributed to the result analysis and provided guidance. All authors contributed to the writing of the manuscript. A. Safa acted as Principal Investigator.}}

\markboth{IEEE Robotics and Automation Letters. Preprint version.}%
{\textsc{Spindle}: Execution-Shortfall Correction in VLA Action Chunks}

\maketitle

\begin{abstract}
Action-chunked vision-language-action (VLA) policies amortise high inference costs by executing predicted trajectory chunks open-loop. On resource-constrained, low-cost manipulators, this open-loop execution renders the robot highly vulnerable to joint friction, tracking lag, and actuator command attenuation. We present \textbf{SPINDLE} (\emph{Shortfall Proprioceptive Inversion for Next-step Disturbance Local Execution}), a fine-tuning-free execution-layer corrector for frozen VLAs. It pairs an offline-calibrated, locally linear command-to-motion model with a 0.18M-parameter learned residual of the nominal dynamics, both fitted on clean rollouts. At every control step, it compares predicted and measured proprioceptive states, inverts the shortfall through the calibrated gains, and adds a bounded correction to the next action, without re-querying the VLA. On 300 held-out LIBERO-Object contexts with a 50-step open-loop horizon, SPINDLE improves task success by 24.3--32.3 percentage points (pp) under actuator attenuation, by 28.0--33.3~pp under transient all-joint Coulomb friction, and by 17.7--28.3~pp under sustained friction; a single calibration fitted on SmolVLA-T rollouts is reused unchanged for two other different policies. In 240 physical trials on a 6-DOF myCobot 320, SPINDLE elevates object placement success rate  by 6.7 to 13.3 pp across both native and zero-shot VLA architectures while reducing query frequency, operating at only 0.4 ms of CPU time per step. Project website: 
\texttt{https://farida-ali.github.io/spindle/}

\end{abstract}


% \begin{figure}[t]
% \centering
% \input{figs/fig0_teaser}
% \caption{\textsc{Spindle} closes the loop that action chunking opens. (a) A
% frozen VLA commits a 50-step chunk; if the arm fails to realize a command,
% nothing at the policy level notices. (b) With the same single query,
% \textsc{Spindle} predicts at every step, after a short warm-up, the motion the
% command should produce (1), measures the state (3), and re-issues the bounded
% shortfall on the next action (4 to 6; stages of Fig.~\ref{fig:overview}).
% Frames and trace: one held-out context, friction on all seven joints at steps
% 15 to 19. Bottom: paired success gain over Base (Table~\ref{tab:main},
% Fig.~\ref{fig:faults}b); n.s.: interval includes zero.}
% \label{fig:teaser}
% \end{figure}

% \begin{figure*}[t]
% \centering\includegraphics[width=\textwidth]{figs/graph_abs.png}
% \caption{ \textbf{Overview of the SPINDLE closed-loop execution correction architecture.} The pre-trained VLA policy generates an open-loop action chunk $\mathbf{A}_t$. During chunk execution, \emph{SPINDLE} predicts
% the physical outcome expected from each committed command, compares this
% prediction with the measured robot state, and uses the resulting execution
% shortfall to correct the next action in the same chunk.  . As a result, SPINDLE reduces execution errors and improves task performance across diverse VLA policies in both simulation and deployment on a real myCobot 320 robot arm.}
% \label{fig:overview}
% \end{figure*}

% Section I: Introduction.

\section{INTRODUCTION}

\IEEEPARstart{V}{ision-Language-Action} (VLA) models have become a prominent approach to robotic manipulation, enabling language-conditioned control by unifying visual perception, instruction understanding, and action generation in a single end-to-end architecture ~\cite{black2024pi0,kim2025oft,shukor2025smolvla}. To balance the substantial inference latency of large transformer and diffusion backbones with continuous trajectory generation, many modern VLAs adopt \emph{action chunking}~\cite{zhao2023act,chi2023diffusion,black2024pi0,shukor2025smolvla}. Rather than querying the base policy at every control step, the model predicts a sequence of actions executed sequentially over a fixed horizon $H$. This temporal amortization reduces the number of policy queries, but it also means that \emph{between queries, the policy does not observe whether its committed actions are physically realized}; at the policy level, execution is open-loop ~\cite{chopra2026everydayvla,zhao2025vlarail}.

The consequences of open-loop execution are especially pronounced
on resource-constrained, low-cost robotic platforms
~\cite{deisenroth2011lowcost,murali2019pyrobot,din2025vision}.
Joint stiction, Coulomb friction, actuator command attenuation,
and unmodeled link dynamics can make the realized motion fall
short of the commanded motion
~\cite{lu2026would,li2026revip}.
Joint-level faults such as increased friction have been shown
to substantially reduce VLA success even when the commanded
configurations remain physically reachable
~\cite{jo2026jparc}.
In long-horizon manipulation, such shortfalls can compound
into grasp failures or false-completion states in which
execution drifts without the policy's awareness
~\cite{li2026revip,liu2026palm,dalal2025local}.

\begin{figure}[t]
\centering
\includegraphics[width=\linewidth]{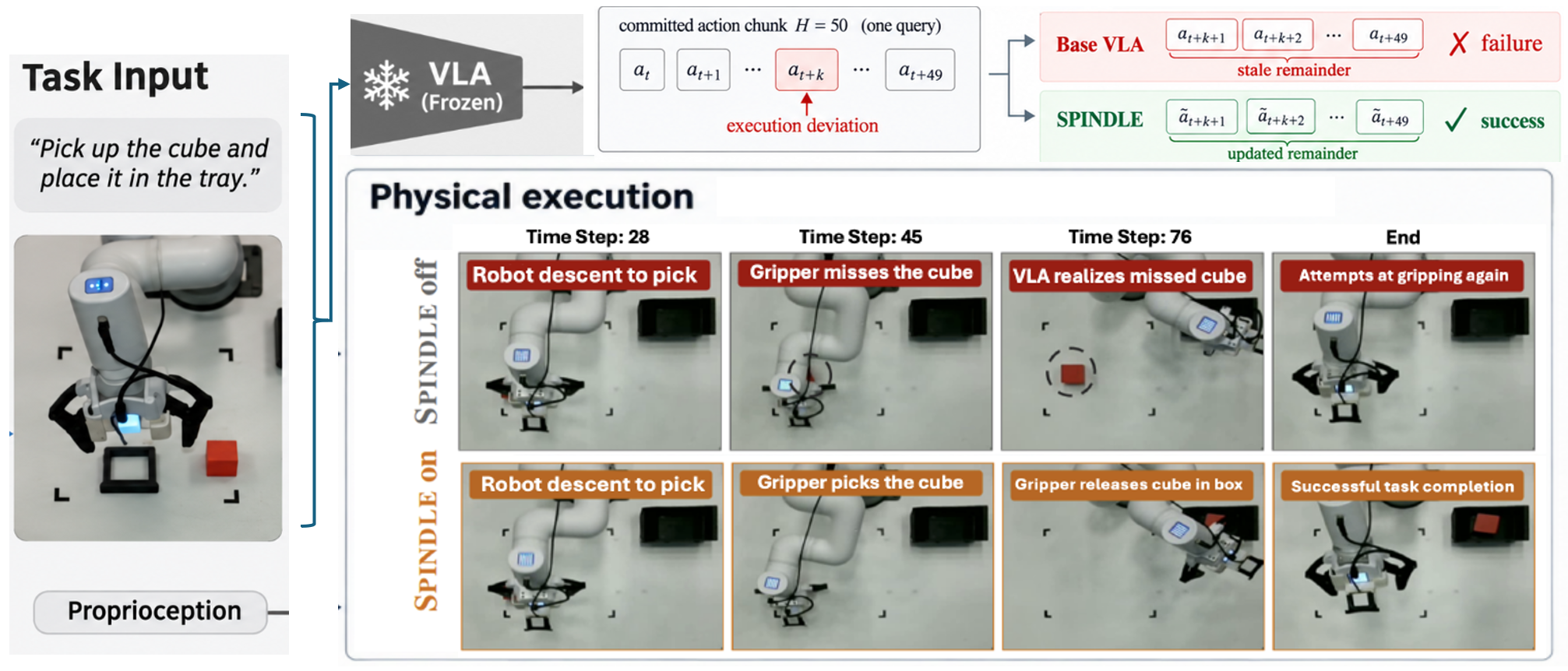}
\caption{\textbf{Overview of \spindle{} for execution-time correction of action-chunked VLAs.} A frozen VLA commits an action chunk for open-loop execution. Following an execution deviation, the uncorrected policy continues with the remaining nominal actions, whereas \spindle{} uses proprioceptive feedback to correct the remaining execution within the same committed chunk without re-querying the VLA. Bottom: matched myCobot 320 rollouts; without SPINDLE, drift leads to a missed grasp and a retry; with it, the grasp and placement succeed.}
\label{fig:1}
\end{figure}
To restore execution fidelity, existing approaches have explored several technical interventions: fine-tuning the base VLA on platform-specific demonstration data, learning auxiliary corrective policies
% ~\cite{yang2026fpc, kim2026objectcentric, liu2025bid}
\cite{yang2026fpc, song2026geobridge, jo2026jparc, sendai2025a2c2, ankile2025resip, kim2026objectcentric}, 
% [20]--[21]
dynamically resampling action chunks 
% \cite{black2025rtc, zhang2026ppc}
\cite{zhang2026ppc, physvla2026},
% [22], [23],
or training secondary monitoring heads to trigger replanning 
% [24]
% \cite{physvla2026}
\cite{zhang2026ppc, physvla2026}. Although capable, these approaches introduce substantial computation, demand extensive hardware-specific data, or risk catastrophic forgetting of pre-trained VLA generalization. Furthermore, secondary neural correctors often policy-dependent and introduce non-trivial computational latency, limiting their utility on embedded platforms with constrained compute budgets 
% [2], [25]
~\cite{kim2025oft}.
% ,mohsen2026mamba}.

To address these limitations, we bridge classical disturbance observer (DOB) principles with modern learning-based dynamics and introduce \spindle{} (\emph{Shortfall Proprioceptive Inversion for Next-step Disturbance Local Execution}), a modular, fine-tuning-free framework that restores physical execution fidelity downstream of frozen VLAs (Fig.~\ref{fig:1}). Rather than relying on rigid analytical plant models that struggle with unmodeled actuator non-linearities, or unconstrained end-to-end correctors that require massive fault-labeled datasets, \spindle{} integrates a lightweight, learned residual forward dynamics predictor with local command-space inversion. Operating as a discrete-time execution monitor at the robot's low-level control rate, \spindle{} pairs an offline-calibrated actuation baseline with a 0.18M-parameter residual MLP to forecast the nominal physical outcome of each committed action. At each step within an active chunk, \spindle{} compares this nominal prediction against real-time proprioception to isolate the true execution shortfall, and causally inverts it into the remaining actions of the committed chunk without triggering high-level policy re-queries.

Because \spindle{} leaves the VLA weights and query schedule unchanged, it corrects execution drift without altering the base policy. The contributions of this letter are:

\begin{enumerate}
    \item \textbf{Execution-layer correction for frozen action-chunked VLAs:} \spindle{} closes the loop on physical execution within each committed chunk using proprioception alone, operating on the normalized action space at the control rate, with no base-model fine-tuning, architectural change, access to internal policy features, or additional policy queries.

    \item \textbf{Hybrid nominal dynamics with algebraic inversion:} An offline-calibrated, per-axis linear actuation model is paired with a learned residual (0.18M parameters) trained only on clean rollouts, so no fault-labeled data are needed. The residual captures state- and history-dependent effects, while the calibrated gains make the shortfall-to-command inversion an element-wise division.

    \item \textbf{Cross-backbone simulation and hardware evaluation:} On 10 LIBERO-Object tasks (300 held-out contexts) with three open-weight VLAs of different architectures and scales, SmolVLA-T~\cite{shukor2025smolvla}, SmolVLA-M~\cite{mohsen2026mamba}, and $\pi_{0.5}$~\cite{intelligence2025pi05}, \spindle{} improves success at $N=50$ by 24.3--32.3~pp under actuator attenuation and by 28.0--33.3~pp under transient joint friction, using one calibration fitted on SmolVLA-T. On SmolVLA-T, it outperforms two recent execution-time correctors, A2C2 and VLA-Corrector, under every tested disturbance. On a 6-DOF myCobot 320 (240 trials), placement success pooled over two backbones rises from 45.0\% to 55.0\%, with fewer VLA queries per success and 0.4~ms of CPU time per control step.
\end{enumerate}

\section{RELATED WORK}
\label{sec:related_work}

Action chunking amortizes the inference cost of diffusion- and flow-based VLAs~\cite{black2024pi0,kim2025oft,shukor2025smolvla}, but the resulting open-loop execution interval leaves low-cost manipulators exposed to tracking lag, unmodeled joint friction, and command undershooting~\cite{jo2026jparc,lu2026would,Kotov2026Backlash}. We group prior work on this execution gap into three lines, summarized in Table~\ref{tab:related_methods}, and then relate \spindle{} to classical disturbance observers.

\begin{table*}[t]
\centering
\caption{Comparison of execution-time correction approaches for frozen action-chunked VLA policies.}
\label{tab:related_methods}

\renewcommand{\arraystretch}{1.1}
\setlength{\tabcolsep}{5pt}

\begin{tabular*}{\textwidth}{@{\extracolsep{\fill}}lccccc@{}}
\hline
\textbf{Method} & \textbf{Corrector / Size} & \textbf{Training Source} & \textbf{Cross-Policy$^*$} & \textbf{VLA Query} & \textbf{Feedback Modality} \\
\hline
ResiP~\cite{ankile2025resip} & Residual policy & Online RL & No & Unchanged & System state \\
A2C2~\cite{sendai2025a2c2} & Residual head / 32M & Demonstrations & No$^\dagger$ & Unchanged & Vision + policy feats. \\
PPC~\cite{zhang2026ppc} & Closed form / 0 & None (training-free) & Yes & Unchanged & Target dynamics \\
PhysVLA~\cite{physvla2026} & State machine / 0 & None (training-free) & Yes & Unchanged & Robot state \\
VLA-Corrector~\cite{zju2026vlacorrector} & Latent model / $\sim$40M & Demonstrations & Yes$^\ddagger$ & Event-triggered & Visual latents \\
J-PARC~\cite{jo2026jparc} & Fault-cond. residual / 2.2M & Faulty + fault-free rollouts & No & Unchanged & Joint dynamics \\
\hline
\textbf{\spindle{} (Ours)} & \textbf{Hybrid dyn. / 0.18M} & \textbf{Clean rollouts} & \textbf{Yes} & \textbf{Unchanged} & \textbf{Proprioception} \\
\hline
\end{tabular*}

\vspace{1mm}
\begin{minipage}{\textwidth}
\footnotesize
$^*$Cross-policy indicates applicability across different frozen VLA backbones without retraining the corrector module.
$^\dagger$A2C2 requires direct access to intermediate feature representations of the specific base VLA.
$^\ddagger$VLA-Corrector can wrap different frozen backbones, but requires training an external $\sim$40M visual dynamics model and interrupts the scheduled query interval.

\end{minipage}
\vspace{-2mm}
\end{table*}

% \subsection{Corrective Frameworks for Action-Chunked Policies}

\subsection{Policy Adaptation and Learned Residuals:} Several methods adapt the policy or train auxiliary modules to counteract physical discrepancies~\cite{song2026geobridge,jo2026jparc,sendai2025a2c2,ankile2025resip,kim2026objectcentric,arslan2026health}. Residual RL~\cite{ankile2025resip,kim2026objectcentric} requires reward design and interactive training. Malfunction-aware VLAs condition the policy on an explicit per-joint health vector, which requires fine-tuning on teleoperated malfunction data~\cite{arslan2026health}. Closest to our setting, J-PARC~\cite{jo2026jparc} shows that joint-level faults, including increased friction, degrade frozen VLAs, and trains a residual calibrator conditioned on a fault regime inferred from recent joint dynamics; its supervision comes from comparing faulty rollouts with fault-free references. A2C2~\cite{sendai2025a2c2} trains a residual head on the latest observation, chunk progress, and intermediate VLA features. In contrast, \spindle{} never observes a fault during training: its learned module is a compact nominal-dynamics predictor (0.18M parameters) fitted on clean rollouts, and corrections are derived algebraically from measured proprioceptive shortfalls. 

% Several methods augment or modify policy parameters to counteract physical discrepancies~\cite{song2025pdvla, jo2026jparc, sendai2025a2c2, ankile2025resip, kim2026objectcentric}. Residual reinforcement learning trains a secondary policy to perturb nominal actions~\cite{ankile2025resip, kim2026objectcentric}, but requires external reward engineering and extensive interactive training loops. Other works leverage execution residuals to update low-rank adaptation (LoRA) weights online; however, such adaptations influence only future action chunks rather than the active, committed trajectory. A2C2~\cite{sendai2025a2c2} preserves the VLA backbone but trains an auxiliary 32M-parameter residual head conditioned on visual observations, raw actions, chunk progress, and intermediate VLA feature representations. In contrast, \spindle{} does not modify VLA weights or rely on internal network features; its learned module is a compact 0.18M-parameter nominal dynamics predictor trained strictly on clean rollouts, with corrective commands derived algebraically from observed proprioceptive shortfalls.

\subsection{ Chunk Replanning and Asynchronous Generation:}

A second class of methods alters how action chunks are generated or refreshed. Parallel decoding accelerates chunk generation~\cite{song2025pdvla}. Bidirectional Decoding (BID)~\cite{liu2025bid} samples multiple candidate trajectories at inference time and selects chunks based on temporal consistency, while Real-Time Chunking (RTC)~\cite{black2025rtc} asynchronously synthesizes future actions during current-chunk execution. VLA-Corrector~\cite{zju2026vlacorrector} adopts an event-triggered recovery approach, employing a $\sim$40M-parameter latent dynamics model to detect visual prediction divergence and truncate the action queue to re-query the VLA with corrective guidance. Recovery therefore relies on additional VLA queries and alters the query schedule. \spindle{} preserves the committed chunk and the original VLA query schedule, executing local corrections at the robot control frequency.

% A second class of methods alters how action chunks are generated or refreshed. Bidirectional Decoding (BID)~\cite{liu2025bid} samples multiple candidate trajectories at inference time and selects chunks based on temporal consistency, while Real-Time Chunking (RTC)~\cite{black2025rtc} asynchronously synthesizes future actions during current-chunk execution. VLA-Corrector~\cite{zju2026vlacorrector} adopts an event-triggered recovery approach, employing a $\sim$40M-parameter latent dynamics model to detect visual prediction divergence and truncate the action queue to re-query the VLA with corrective guidance. However, this recovery mechanism alters the high-level query schedule and increases inference overhead. \spindle{} preserves the committed chunk and the original VLA query schedule, executing local corrections at the robot control frequency.

\subsection{Execution-Time Kinematic Shaping}

Training-free wrappers adjust actions at inference time without altering model weights. Pace-and-Path Correction (PPC)~\cite{zhang2026ppc} dynamically scales trajectory progress and injects orthogonal path offsets in response to external target movement. PhysVLA~\cite{physvla2026} applies a phase-aware state machine with an Euler--Lagrange consistency filter to eliminate dynamically infeasible actions. These methods enforce geometric or dynamic consistency of the commanded actions, but they do not model low-level tracking errors such as actuator undershooting or Coulomb joint friction. \spindle{} targets this complementary failure mode with control-rate proprioceptive feedback.

\subsection{Disturbance Observers and Inversion Methods}

DOBs estimate unmodeled plant dynamics and external disturbances by comparing commanded inputs against measured states to cancel tracking errors at the low-level loop~\cite{ohnishi1987,chen2016dobc}. DOBs are standard in motion control~\cite{chen2016dobc} and have been used to transfer neural driving policies across vehicle dynamics~\cite{tang2020dob}. Applying them to VLAs is not straightforward: VLAs output normalized kinematic actions rather than joint torques, so a conventional inverse-dynamics plant model is not directly available. \spindle{} bridges this gap by decoupling disturbance rejection into a calibrated linear command-to-motion baseline and a lightweight learned residual dynamics predictor, mapping isolated execution shortfalls back into normalized command corrections.

\section{Method}
\label{sec:methods}

\begin{figure*}[t]
    \centering
    \includegraphics[width=0.77\textwidth]{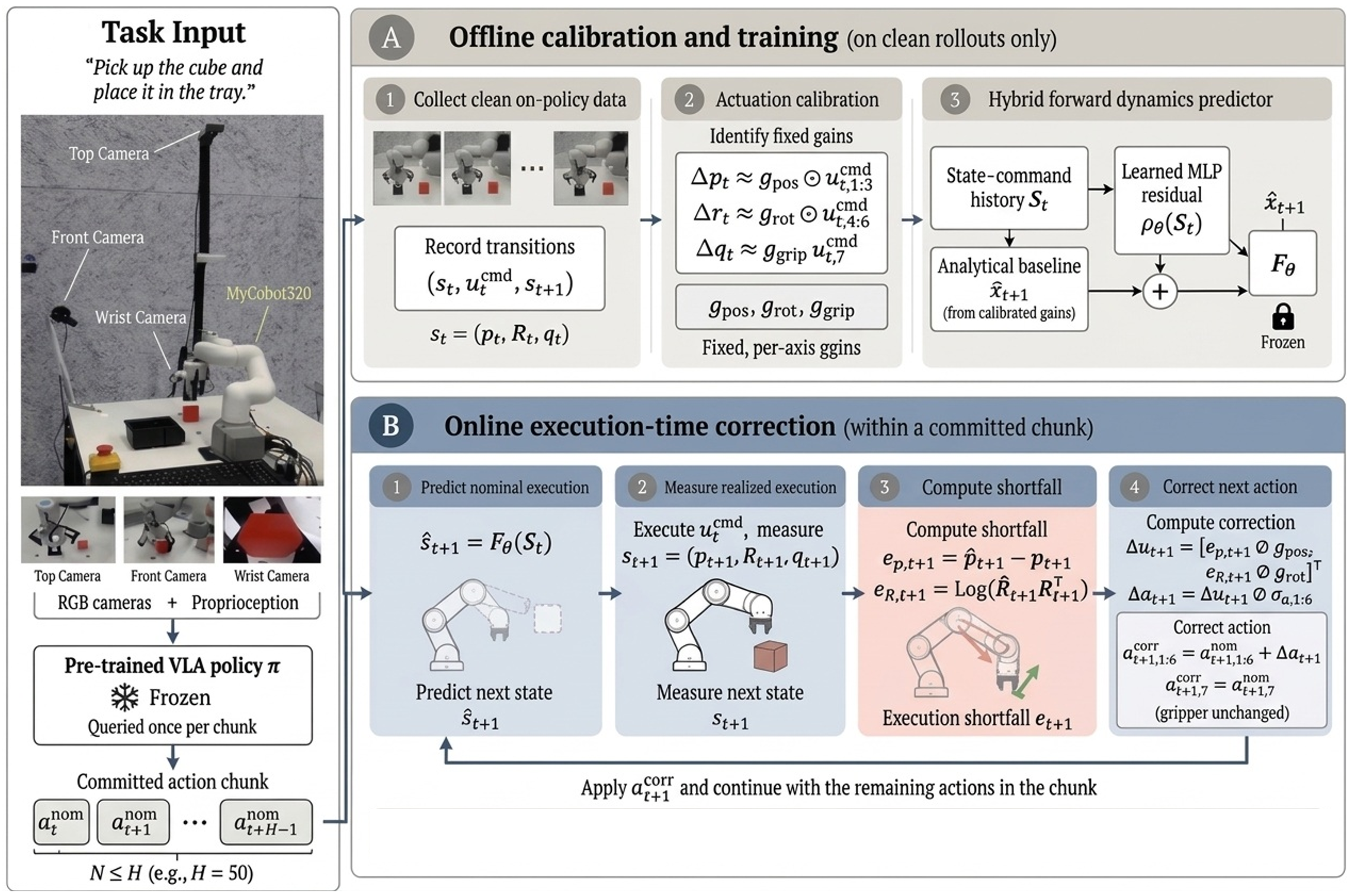}
% \caption{\textbf{Overview of the \spindle{} framework.} Offline, clean on-policy rollouts are used to calibrate the command-to-motion relationship and train the hybrid forward dynamics predictor $F_\theta$, which combines an analytical next-state estimate with a learned residual. During online execution, the frozen VLA provides a nominal action chunk, and \spindle{} compares the predicted and realized robot states after each executed action. The resulting execution shortfall is mapped back to command space through the calibrated actuation model and applied to the next unexecuted action within the active chunk.}

\caption{\textbf{Overview of the \spindle{} execution-time correction framework.}
\textit{Left (Physical setup and policy interface):} The myCobot~320 workspace uses top, front, and wrist RGB views together with robot proprioception; the frozen VLA is queried once per execution interval to produce a committed action chunk.
\textit{(A) Offline calibration and training:} Clean on-policy rollouts are used to calibrate the local command-to-motion response and train the learned residual of the hybrid forward dynamics predictor.
\textit{(B) Online execution-time correction:} During chunk execution, \spindle{} predicts nominal motion, compares the prediction with the measured robot state, estimates the resulting execution shortfall, and corrects the next unexecuted action. The correction operates within the committed chunk without corrective VLA queries or VLA weight updates.}

     \label{fig:spindle_method_workflow}
\end{figure*}

\spindle{} provides execution-time correction for a frozen, action-chunked VLA by closing the physical command--realization loop within each committed action chunk. For every executed command, it predicts the expected next robot state, compares this prediction with proprioceptive measurements, and uses the resulting execution shortfall to correct the next unexecuted action. The VLA weights and query schedule remain unchanged. Model-based disturbance compensation typically relies on an explicit command-to-motion model that can be inverted to recover the control adjustment required to counteract an observed execution error. For low-cost manipulators, however, the physical response to commanded actions can exhibit nonlinear friction and drivetrain uncertainty that are difficult to capture accurately with a fixed analytical model~\cite{dou2026neuralactuator}. \spindle{} therefore represents nominal execution using a calibrated local command-to-motion baseline together with a lightweight data-driven residual.

As illustrated in Fig.~\ref{fig:spindle_method_workflow}, \spindle{} comprises four components. \textit{Actuation Calibration} first estimates the local command-to-motion relationship from clean on-policy rollouts (Sec.~\ref{sec:actuation_calibration}). \textit{Proprioceptive Forward Dynamics Prediction} then combines the resulting analytical next-state estimate with a learned residual to predict nominal execution (Sec.~\ref{sec:hybrid_dynamics}). After each command, \textit{Execution Shortfall Correction} compares the predicted and measured states to compute the execution shortfall and maps it back to command space through the calibrated actuation model (Sec.~\ref{sec:shortfall_correction}). Finally, the \textit{Online Execution Loop} applies the resulting correction causally to the next unexecuted action within the same chunk (Sec.~\ref{sec:online_pipeline}).

\subsection{Problem Formulation}
\label{sec:problem_formulation}

We consider a robotic manipulator equipped with a parallel-jaw gripper and
controlled at a fixed control rate through a low-level controller. At control step $t$, the robot executes one command; the  frozen VLA policy $\pi$ is queried every N steps and receives a multimodal observation
$o_t$, comprising  camera views and robot proprioception, together with a
natural-language instruction $\ell$. The measured physical state is denoted
by $s_t=(p_t,R_t,q_t)$, where $p_t\in\mathbb{R}^3$ is the end-effector
position, $R_t\in\mathrm{SO}(3)$ is its orientation in the robot base frame,
and $q_t\in\mathbb{R}^2$ contains the measured gripper-finger positions.

To mitigate inference latency, the policy predicts an action chunk of $H$
future nominal actions:
\begin{equation}
A_t = \pi(o_t,\ell)
=
\left[
a_t^{\mathrm{nom}},
a_{t+1}^{\mathrm{nom}},
\dots,
a_{t+H-1}^{\mathrm{nom}}
\right],
\label{eq:action_chunk}
\end{equation}
where
$a_t^{\mathrm{nom}}
=
[a_{t,1:3}^{\mathrm{nom}},
a_{t,4:6}^{\mathrm{nom}},
a_{t,7}^{\mathrm{nom}}]^\top
\in\mathbb{R}^7$
contains three translational, three rotational, and one gripper component.
During deployment, the robot commits to executing $N\leq H$ actions before
querying $\pi$ again.

Each policy action is mapped by a policy-specific affine postprocessor
$\phi(\cdot)$ to the controller-space command
\begin{equation}
    u_t^{\mathrm{cmd}} = \phi(a_t^{\mathrm{nom}}) \in \mathbb{R}^7,
\end{equation}
where $u_{t,1:3}^{\mathrm{cmd}}$ specifies Cartesian translation,
$u_{t,4:6}^{\mathrm{cmd}}$ specifies end-effector rotation, and
$u_{t,7}^{\mathrm{cmd}}$ controls the gripper.
 The low-level controller maps
these commands to the corresponding robot actuation. We distinguish the
policy action $a_t^{nom}$ from $u_t^{\mathrm{cmd}}$ because \spindle{} models the
physical response to the command presented to the low-level controller.

Under nominal execution, applying $u_t^{\mathrm{cmd}}$ from state $s_t$
produces an expected one-step physical response. Hardware non-idealities may
instead introduce an unknown execution disturbance $d_t$, yielding
\begin{equation}
s_{t+1}
=
\mathcal{P}(s_t,u_t^{\mathrm{cmd}},d_t),
\label{eq:real_dynamics}
\end{equation}
where $\mathcal{P}$ represents the unknown coupled robot--controller dynamics
and $d_t$ denotes the aggregate effect of unmodeled execution disturbances.
\textcolor{black}{\spindle{} does not estimate $d_t$ explicitly; instead, it
infers its effect from the discrepancy between the predicted nominal response
and the subsequently measured state.}

The objective of \spindle{} is to estimate the discrepancy between the
expected nominal response and the realized state after each executed command
and causally map it to a spatial correction
$\Delta a_{t+1}\in\mathbb{R}^{6}$ for the next unexecuted action:
\begin{align}
a_{t+1,1:6}^{\mathrm{corr}}
&=
a_{t+1,1:6}^{\mathrm{nom}}+\Delta a_{t+1},
\rev{\qquad}
a_{t+1,7}^{\mathrm{corr}}
= a_{t+1,7}^{\mathrm{nom}}.
\label{eq:action_update_spatial}
\end{align}

\subsection{Actuation Calibration and Nominal Baseline}
\label{sec:actuation_calibration}

We first construct \textcolor{black}{a local linear} command-to-motion baseline
that provides both an analytical next-state forecast and an invertible mapping
for subsequent shortfall compensation. To identify these baseline gains, we
collect unperturbed on-policy transitions from the frozen base policy
\textcolor{black}{under clean execution}, with \spindle{} disabled and no
injected execution disturbances. For each transition step $t$, we record the
issued command $u_t^{\mathrm{cmd}}\in\mathbb{R}^{7}$ alongside the measured
one-step translational, rotational, and gripper increments:
\begin{equation}
\begin{aligned}
\Delta p_t &= p_{t+1}-p_t, \qquad
\Delta r_t = \mathrm{Log}\!\left(R_{t+1}R_t^\top\right),\\
\Delta q_t &= q_{t+1}-q_t ,
\end{aligned}
\label{eq:motion_increments}
\end{equation}
where $\mathrm{Log}:\mathrm{SO}(3)\rightarrow\mathbb{R}^{3}$ maps the
relative rotation to its axis-angle representation, taken here in the robot
base frame ~\cite{barfoot2017state}. We identify per-axis nominal actuation gains by ordinary least
squares:
\begin{equation}
\begin{aligned}
\Delta p_t
&\approx
g_{\mathrm{pos}}\odot u_{t,1:3}^{\mathrm{cmd}},\\
\Delta r_t
&\approx
g_{\mathrm{rot}}\odot u_{t,4:6}^{\mathrm{cmd}}, \\
\Delta q_t
&\approx
g_{\mathrm{grip}}\,u_{t,7}^{\mathrm{cmd}},
\end{aligned}
\label{eq:calibration_gains}
\end{equation}
where $g_{\mathrm{pos}},g_{\mathrm{rot}}\in\mathbb{R}^{3}$ denote the fitted
translational and rotational command gains,
$g_{\mathrm{grip}}\in\mathbb{R}^{2}$ denotes the gripper gain, and $\odot$
represents element-wise multiplication. We retain diagonal gains as a local approximation because the controller
interface is axis-parameterized and the measured cross-axis coupling is
limited. A full $3\times3$ refit reduced the held-out translational error only
from 1.41 to 1.33\,mm, while retaining the diagonal model enables direct
element-wise inversion. 

% Remaining coupled and state-dependent effects are
% captured by the learned residual in Sec.~\ref{sec:hybrid_dynamics}.

% We retain diagonal gains because the controller interface is
% axis-parameterized, mapping each command component to a goal offset along
% the corresponding base-frame axis, and empirical cross-axis coupling is
% small: 7\% of the translation block's norm and 9\% of the rotation
% block's lie off the diagonal (scale-free), and refitting full $3{\times}3$
% blocks lowers the held-out one-step error only from 1.41 to 1.33\,mm, whereas
% the learned residual reaches 0.66\,mm. This keeps the inversion
% in~\eqref{eq:command_inversion} a direct element-wise division while residual coupled and state-dependent dynamics are absorbed by the
% learned predictor in Sec.~\ref{sec:hybrid_dynamics}.

\subsection{Hybrid Proprioceptive Forward Dynamics Prediction}
\label{sec:hybrid_dynamics}

The calibrated model captures the dominant command-to-motion response, but
nominal execution also depends on the robot state and recent motion history.
\spindle{} therefore augments the analytical prediction with a lightweight
learned residual trained exclusively on clean on-policy rollouts.

Let $\bar{x}_{t+1}$ denote the analytical next-state estimate obtained by
propagating the current state using the calibrated gains in
Eq.~\eqref{eq:calibration_gains}. To capture dynamics not represented by this
local model, a lightweight learned residual network $\rho_\theta$ (0.18M parameters) processes the most recent
$K=10$ state--command pairs,
$S_t=\{(x_i,u_i^{\mathrm{cmd}})\}_{i=t-K+1}^{t}$,
and predicts a residual correction. The resulting hybrid prediction is
\begin{equation}
\hat{x}_{t+1}
=
F_\theta(S_t)
=
\Pi\!\left[
\bar{x}_{t+1}
+
\sigma_x\odot\rho_\theta(S_t)
\right],
\label{eq:hybrid_prediction}
\end{equation}
where
$x_t=[p_t^\top,r_6(R_t)^\top,q_t^\top]^\top\in\mathbb{R}^{11}$
is the predictor state representation, with $r_6(\cdot)$ denoting the
continuous 6-D orientation representation~\cite{zhou2019rotation}.
Here, $\sigma_x$ restores the residual to the state scale, while $\Pi$
ensures a valid rotation representation after residual addition. Crucially,
$\rho_\theta$ is trained exclusively on clean on-policy rollouts and requires
no fault-labeled training data. Both the calibrated gains and $\rho_\theta$
remain frozen during deployment. Thus, $\hat{x}_{t+1}$ provides the nominal
execution reference against which the subsequently measured state is compared
to quantify the execution shortfall.

\subsection{Execution Shortfall Estimation and Causal Command Inversion}
\label{sec:shortfall_correction}

Following execution of $u_t^{\mathrm{cmd}}$, proprioception provides the
measured physical state
$s_{t+1}=(p_{t+1},R_{t+1},q_{t+1})$. \textcolor{black}{From the predicted
state $\hat{x}_{t+1}$, we recover the corresponding position and orientation
$(\hat{p}_{t+1},\hat{R}_{t+1})$.}
\spindle{} quantifies the spatial execution shortfall by comparing this
prediction with the realized state:
\begin{equation}
e_{p,t+1}
=
\hat{p}_{t+1}-p_{t+1},
\qquad
e_{R,t+1}
=
\mathrm{Log}\!\left(
\hat{R}_{t+1}R_{t+1}^{\top}
\right),
\label{eq:shortfall_computation}
\end{equation}
where $e_{p,t+1},e_{R,t+1}\in\mathbb{R}^{3}$ represent translational and
rotational deviations from the expected nominal response.
\textcolor{black}{The gripper state is also predicted by the forward model but
is not corrected; \spindle{} modifies only the six spatial action dimensions.}

To map these operational-space shortfalls into controller-command
compensations, \spindle{} applies the inverse of the calibrated spatial
actuation model~\eqref{eq:calibration_gains}:
\begin{equation}
\Delta u_{t+1}
=
\begin{bmatrix}
e_{p,t+1}\oslash g_{\mathrm{pos}}\\
e_{R,t+1}\oslash g_{\mathrm{rot}},
\end{bmatrix}
\in\mathbb{R}^{6},
\label{eq:command_inversion}
\end{equation}
where $\oslash$ denotes element-wise division.
Under the local diagonal approximation in~\eqref{eq:calibration_gains}, the
element-wise division directly maps the unrealized translational and rotational motion back into the corresponding controller-command units. Then, because the policy postprocessor is affine, this command-space compensation
is mapped back to the spatial policy-action coordinates through the action
scales $\sigma_{a,1:6}$:
\begin{equation}
\Delta a_{t+1}
=
\Delta u_{t+1}\oslash\sigma_{a,1:6}
\in\mathbb{R}^{6}.
\label{eq:action_space_projection}
\end{equation}
%
% \textcolor{red}{
% The affine offset cancels for a command difference. Since the shortfall is
% observed only after step $t$ has been executed, \spindle{} applies
% $\Delta a_{t+1}$ causally to the next unexecuted action according
% to~\eqref{eq:action_update_spatial}, while preserving the nominal gripper
% action.}

\subsection{Online Execution Pipeline}
\label{sec:online_pipeline}

During each committed VLA chunk, \spindle{} operates synchronously at the
control rate. For the current action $a_t^{\mathrm{corr}}$, the corresponding
command $u_t^{\mathrm{cmd}}$ and measured state complete the latest entry in
$S_t$, from which the hybrid predictor estimates the nominal next state
$\hat{x}_{t+1}$. After execution, the measured state $s_{t+1}$ yields the
execution shortfall, which is mapped through~\eqref{eq:command_inversion}
and~\eqref{eq:action_space_projection} to a correction for the next
unexecuted action. This cycle continues within the active chunk without
regenerating it, modifying the frozen VLA, or issuing an additional
corrective query.

\begin{figure*}[t]
\centering
\includegraphics[width=\textwidth]{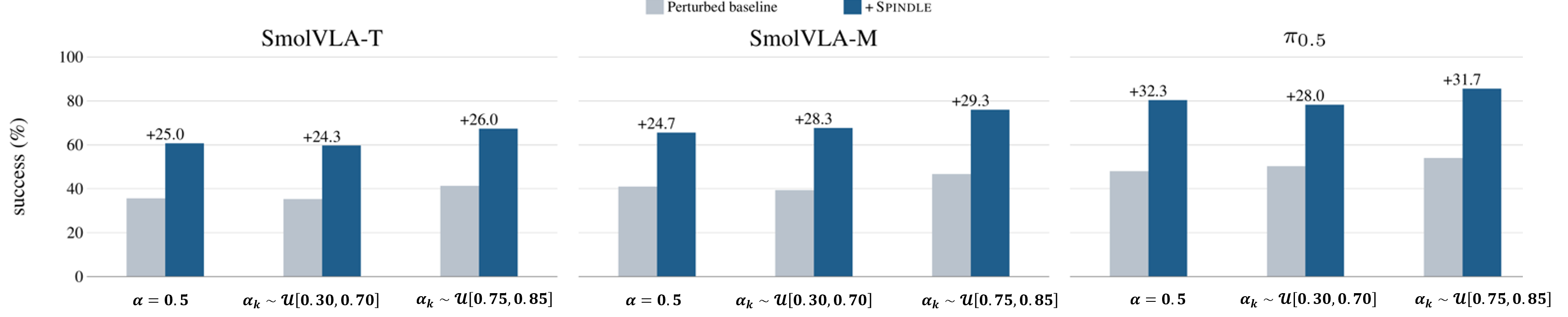}
\caption{\textbf{\spindle{} under simulated actuator undershooting on LIBERO-Object.} Task success over 300 held-out contexts per bar; labels give the paired gain (pp) of SPINDLE over the perturbed Base.}
\label{fig:faults_att}
\end{figure*}

\section{Experimental Results}
\label{sec:exp}

\subsection{Experimental Setup}
\label{sec:setup}

\textcolor{black}{To systematically assess the corrective capabilities of} \spindle{}, we conduct extensive experimentation across both \textit{simulated} benchmark environments and \textit{physical} hardware deployments.

\subsubsection{Policies and Simulation Benchmark} 
\label{simu_setup}
We evaluate \spindle{} on three VLA architectures: SmolVLA-T \cite{shukor2025smolvla}, which uses Transformer self-attention in its action expert; SmolVLA-M~\cite{mohsen2026mamba}, which replaces these layers with Mamba blocks; and $\pi_{0.5}$~\cite{intelligence2025pi05}. Here, SmolVLA-T serves as the primary backbone, while SmolVLA-M and $\pi_{0.5}$ evaluate zero-shot cross-architecture generalization. All policies are fine-tuned for 30,000 steps on LIBERO~\cite{liu2023libero} and predict action chunks of length $H=50$.

Simulation experiments are conducted across ten tasks in the LIBERO-Object benchmark~\cite{liu2023libero} within robosuite~\cite{zhu2020robosuite}, using an operational-space pose controller running at $20\,\text{Hz}$. To prevent data leakage and evaluate cross-context generalization, \textcolor{black}{both} the forward dynamics predictor (\text{Sec.~\ref{sec:hybrid_dynamics}}) and actuation calibration gains (\text{Sec.~\ref{sec:actuation_calibration}}) are fitted exclusively on clean SmolVLA-T rollouts across initial states $0$--$19$ of four tasks. Evaluation uses held-out initial states 20--49 across all ten tasks, yielding 300 contexts per policy.

% \begin{figure}[t]
% \centering
% \includegraphics[width=0.7\linewidth]{figs/setup.pdf}
% \caption{\textbf{Physical \textit{myCobot 320} setup for tabletop manipulation.} \textit{Left:} 6-DOF robot workspace with a parallel-jaw gripper, target cube, tray, and three cameras. \textit{Right:} Multi-view observations provided to the VLA policy from the \textit{top}, \textit{front}, and \textit{wrist} cameras.}
% \label{fig:mycobotsetup}
% \end{figure}

\paragraph{Disturbance Injection.} To rigorously stress-test \spindle{} under simulated execution faults, we introduce two classes of disturbances. \textit{Actuator undershooting} models joint-command attenuation common in budget hardware by scaling all joint commands by a factor $\alpha\in(0,1]$. We consider fixed attenuation ($\alpha=0.5$) and stochastic per-step attenuation, with $\alpha_k\sim\mathcal{U}[0.30,0.70]$ or $\mathcal{U}[0.75,0.85]$, beginning at step 15 of the first action chunk. We additionally inject dry (Coulomb) friction~\cite{armstrong1994survey} downstream of the low-level controller using MuJoCo's~\cite{todorov2012mujoco} \texttt{frictionloss} constraint bounded by $c_j$. This models uncalibrated or worn mechanical joints through a constant resistance torque opposing motion. Joint friction is evaluated under two  regimes: (i) a five-step transient disturbance in the first action chunk, representing brief effects such as stiction or mechanical snags, and (ii) a sustained disturbance from step 15 to episode completion, representing persistent effects such as joint wear or added drag. These regimes assess recovery from transient faults and continuous compensation under persistent resistance, respectively.

\paragraph{Evaluation Protocol} We report task success over 300 held-out LIBERO-Object contexts, with Table~\ref{tab:baseresults} establishing fault-free performance with and without \spindle{}. Under disturbance, we report both the task success rate (\%) and the paired recovery achieved by \spindle{} relative to the corresponding disturbed baseline in percentage points (pp). All comparisons are paired over identical contexts; 95\% confidence intervals are obtained using a task-stratified cluster bootstrap, with exact McNemar tests for paired binary outcomes. We adopt $N = 50$ as the primary evaluation regime because it represents the maximum open-loop execution window, providing the most rigorous test of whether \spindle{} can close the control loop mid-chunk to mitigate physical execution faults. Query efficiency is reported as \emph{VLA queries per success}. We  further ablate the learned residual $\rho_\theta$ from $F_\theta$
while retaining its calibrated analytical component. All methods are evaluated
on identical context--disturbance pairs. Additionally, we examine the success--query trade-off by evaluating $N\in\{10,25,40,50\}$ on initial states 35--49 across all ten tasks, yielding 150 contexts per operating point. 

% \paragraph{Evaluation Protocol} 
% \textcolor{black}{To evaluate policy performance, we report} the mean task success rate across $300$ held-out contexts on LIBERO-Object under an open-loop execution horizon. Table~\ref{tab:baseresults} reports the unperturbed baseline performance of all three VLA architectures, both with and without \spindle{}, establishing the baseline for disturbance testing and assessing its effect on fault-free execution. \textcolor{black}{Accordingly,} our primary evaluation metric is the paired success rate recovery in percentage points ($\text{pp}$) achieved upon enabling \spindle{} under disturbance. \textcolor{red}{All contrasts are paired over contexts, with 95\% intervals from a task-stratified cluster bootstrap and exact McNemar tests. Cost is measured in \emph{queries per success}: the total number of VLA queries over all episodes, failures included, divided by the number of successes.} We adopt $N = 50$ as the primary evaluation regime because it represents the maximum open-loop execution window, providing the most rigorous test of whether \spindle{} can close the control loop mid-chunk to mitigate physical execution faults. \textcolor{black}{Furthermore, to characterize the interaction between open-loop horizon length and closed-loop corrective feedback,} we systematically vary $N \in \{10, 25, 40, 50\}$ across initial states $35$--$49$ ($150$ contexts).

\paragraph{Comparison Methods}
We compare \spindle{} with VLA-Corrector~\cite{zju2026vlacorrector} and
A2C2~\cite{sendai2025a2c2}, using their released implementations adapted to
SmolVLA-T. VLA-Corrector detects visual prediction error and triggers
corrective replanning, whereas A2C2 learns a per-step residual correction
conditioned on the latest observation, the base action, the position within
the active chunk, and features from the base policy. For A2C2, the correction
scale is set to $0.5$ using calibration contexts disjoint from the evaluation
set.

\begin{figure*}[h]
    \centering
    \includegraphics[width=0.485\textwidth]{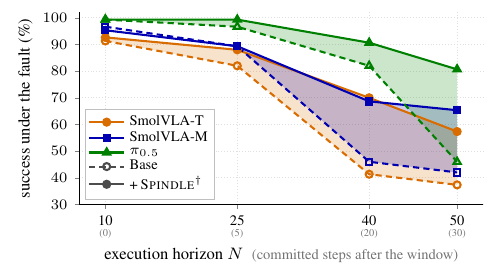}
    \hfill
    \includegraphics[width=0.5\textwidth]{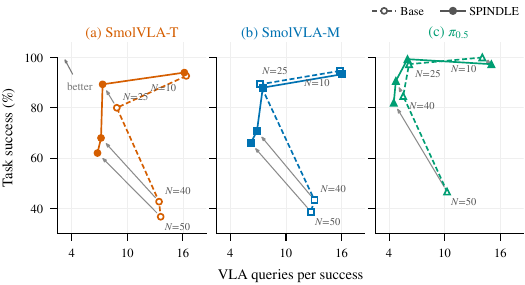}

\caption{Effect of the execution horizon on robustness and VLA-query efficiency under actuator attenuation. \textit{Left:} Task success across execution horizons $N\in\{10,25,40,50\}$ for the perturbed Base (dashed) and +\spindle{} (solid) on the three VLA backbones; values in parentheses give the committed steps that remain after the fault window. \textit{Right:} Corresponding success--query operating points for each backbone; up and to the left is better. Initial states 35--49, 150 contexts per point. }

    \label{fig:dynamicshorizon}
\end{figure*}

\begin{table}[htbp]
\renewcommand{\arraystretch}{1.0}
\caption{Task success (\%) without injected faults at $N=50$ on 300 held-out contexts per policy, without and with \spindle{}. \spindle{}.}
\label{tab:baseresults}
\centering
\begin{tabular}{lccc}

\hline
 & \textbf{SmolVLA-T} & \textbf{SmolVLA-M} & \textbf{$\pi_{0.5}$} \\
\hline
Base              & 65.3 & 73.0 & 92.0 \\
+\spindle{} & 68.0 & 74.0 & 88.7 \\
\hline
\end{tabular}
\end{table}

\subsubsection{Real-World Hardware Deployment}
\label{sec:setup:realworld}
To complement the simulated disturbance evaluation, we further assess \spindle{} under unmodeled hardware non-idealities and physical execution errors on a low-cost robotic platform. Specifically, we deploy \spindle{} on a physical 6-DOF \textit{myCobot 320} manipulator integrated into a custom VLA evaluation environment with a two-finger parallel-jaw gripper and three RGB camera views, as shown in Fig.~\ref{fig:spindle_method_workflow}. The robot is tasked with following the language instruction ``Pick up the cube and put it in the box,'' which requires grasping a red cube and placing it inside a fixed tray. We use SmolVLA-T as the calibration backbone and SmolVLA-M to evaluate zero-shot cross-architecture transfer. We omit the larger$\pi_{0.5}$ to test real-time control on constrained hardware. Both VLAs are fine-tuned on 190 teleoperated demonstrations \footnote{ \url{https://huggingface.co/datasets/telkaffash/mycobot-pickplace-full}} spanning a $5\times5$ workspace grid and additional randomly sampled object locations. Finally, \spindle{}'s actuation gains and $F_\theta$ are fitted using 85 clean SmolVLA-T trajectories and transferred to SmolVLA-M without recalibration. For each backbone, we conduct 60 baseline and 60 \spindle{} trials at 10~Hz. A trial succeeds when the robot grasps the cube and releases it inside the target tray. Because chunks are requested asynchronously, chunks after the first execute a median of 43 actions for SmolVLA-T and 42 for SmolVLA-M before the next chunk becomes active. 

\subsection{Recovery under Simulated Actuator Undershooting}
\label{sec:results:main_shooting}
% First, in order to study the effect of \spindle{} on common actuator undershooting issues found in many low-cost robot manipulators (such as the \textit{myCobot 320} used in this work), Fig. \ref{fig:faults_att} shows the performance of the three VLA architectures under simulated actuator attenuation applied to all robot joints within the LIBERO environment (making the disturbance level highly challenging). To capture diverse disturbance conditions, we evaluate performance under both fixed attenuation  ($\alpha=0.5$) and uniformly distributed random attenuation across two severity levels ($\alpha \sim \mathcal{U}_{0.75,0.85}$ and $\alpha \sim \mathcal{U}_{0.30,0.70}$) during the first 35 execution steps. Fig. \ref{fig:faults_att} remarkably shows that the proposed \spindle{} method \textit{significantly boosts} success rates by $27.7 pp$ on average across models the heavily degraded success rates obtained for all models when affected by the actuator under-shooting non-ideality (see model degradations between Table \ref{tab:baseresults} and the gray bars in Fig. \ref{fig:faults_att}).

Our experimental results show that \spindle{} \textit{substantially improves} VLA robustness to actuator undershooting across all three evaluated backbones and attenuation regimes defined in Section~\ref{simu_setup}. As shown in Fig.~\ref{fig:faults_att}, actuator attenuation heavily degrades the uncorrected success rates relative to the clean baselines in Table~\ref{tab:baseresults}, whereas \spindle{} recovers task success by $27.7~pp$ on average across models. Importantly, the corrector calibrated on SmolVLA-T transfers to SmolVLA-M and $\pi_{0.5}$ without recalibration, indicating that the correction requires neither backbone-specific nor disturbance-specific retuning.

The obvious alternative to within chunk correction is a shorter horizon N with more frequent queries. As shown in Fig.~\ref{fig:dynamicshorizon} (left), frequent VLA inference at $N=10$ already yields high success, leaving limited room for further improvement. However, this robustness comes at a considerably higher inference cost, requiring approximately 14--16 VLA queries per successful episode. As $N$ increases and VLA queries become less frequent, the uncorrected policies degrade sharply, while the benefit of \spindle{} becomes progressively larger. At $N=50$, \spindle{} raises success from 37.3\% to 57.3\% on SmolVLA-T, from 42.0\% to 65.3\% on SmolVLA-M, and from 46.0\% to 80.7\% on $\pi_{0.5}$, corresponding to gains of 20.0, 23.3, and 34.7~pp, respectively. The corresponding success--query operating points in Fig.~\ref{fig:dynamicshorizon} (right) show that \spindle{} recovers performance in this low-query regime without issuing corrective VLA queries. 

% Consequently, \spindle{} shifts the success--query trade-off favorably, recovering robustness at longer horizons while retaining their lower VLA inference cost.

\begin{figure*}[t]
\centering
\includegraphics[width=\textwidth]{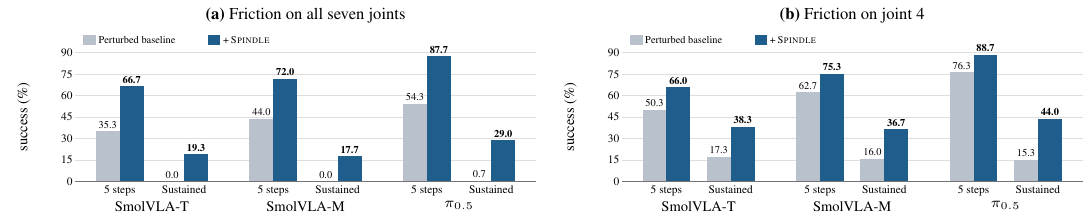}\\
\caption{\textbf{Task success rates under simulated Coulomb joint friction on LIBERO-Object} Performance comparison between perturbed baselines (gray) and +\spindle{} (blue) across SmolVLA-T, SmolVLA-M, and $\pi_{0.5}$ under joint friction regimes. }
\label{fig:faults_fric}
\end{figure*}
% \textcolor{orange}{As the bottleneck that chunk-
% ing was designed to relieve is reducing inference overhead by executing larger chunk sizes to reduce the nuber of VLA queiries}. So Furthermore, Fig.~\ref{fig:dynamicshorizon} illustrates the effect of the
% open-loop execution horizon $N$ on the model success rates, and the gains provided by \spindle{} under simulated actuator undershooting. Across all three VLA architectures, Fig.~\ref{fig:dynamicshorizon} shows a consistent pattern: as the execution horizon increases, the \textit{uncorrected} baseline success rates degrade sharply. At $N=50$, baseline success drops from $66.0\%$ to$37.3\%$ for SmolVLA-T, $72.7\%$ to $42.0\%$ for SmolVLA-M, and $94.7\%$ to
% $46.0\%$ for $\pi_{0.5}$. In contrast, \spindle{} significantly alleviates these degradations, raising success rates to 57.3\% (+20.0\,pp) on SmolVLA-T, 65.3\% (+23.3\,pp) on SmolVLA-M, and 80.7\% (+34.7\,pp) on $\pi_{0.5}$ at $N = 50$. These results establish \spindle{} as an effective execution-time corrector for mitigating performance degradation at long open-loop execution horizons. \textcolor{orange}{These results proves the aim of \spindle{} which targets improving robustness in the low-query regime: as the committed execution horizon grows, it substantially recovers disturbance-induced failures without introducing corrective VLA calls as shown in Fig. X b. As reducing the execusion horizon is advantegous, howerever the multiply inference overhead–the bottleneck that chunk-
% ing was designed to relieve
% }

\subsection{Recovery under  Simulated Joint Friction}
\label{sec:results:arm}

We next evaluate \spindle{} under the joint-friction regimes defined in Section~\ref{simu_setup}, with results for all-joint and Joint-4 friction reported in Fig.~\ref{fig:faults_fric}. Under transient all-joint friction (Fig.~\ref{fig:faults_fric}a), the baseline success rates drop sharply to 35.3\%, 44.0\%, and 54.3\% for SmolVLA-T, SmolVLA-M, and $\pi_{0.5}$, respectively. Integrating \spindle{} decisively recovers performance across all three backbones, raising success rates to 66.7\% (+31.3pp), 72.0\% (+28.0pp), and 87.7\% (+33.3 pp) for the respective backbones. Sustained all-joint friction is considerably more severe, reducing the uncorrected success rates to near zero (0.0\%, 0.0\%, and 0.7\%); nevertheless, \spindle{}  recovers task success to 19.3\%, 17.7\%, and 29.0\%, respectively. A similar trend is observed under Joint-4 friction, where \spindle{} yields mean gains of 13.6~pp under transient friction and 23.5~pp under sustained friction across the three backbones. Together with the actuator-undershooting results in Section~\ref{sec:results:main_shooting}, these findings show that the same calibrated \spindle{} corrector provides consistent recovery across transient and persistent mechanical disturbances without disturbance-specific retuning or backbone-specific recalibration.

\subsection{Effect on Nominal Performance}
We also assess the effect of \spindle{} on nominal policy performance under clean execution. As shown in Table~\ref{tab:baseresults}, \spindle{} preserves and slightly improves success on SmolVLA-T and SmolVLA-M by +2.7 and +1.0~pp, respectively, while $\pi_{0.5}$ shows a modest 4.3~pp decrease under zero-shot transfer. We retain this setting without backbone-specific recalibration to preserve the same transfer protocol throughout the evaluation.

\subsection{Comparison with Existing VLA Correctors}
\begin{table}[t]
\centering
\caption{Comparison of execution-time correctors on
SmolVLA-T and LIBERO-Object ($N=50$; 300 paired contexts).
Entries report success (\%) and VLA queries per success
in parentheses (omitted below five successes). $\Delta_{\mathrm{best}}$ denotes the
success-rate improvement (pp) over the strongest competitor. }
\label{tab:correctors}
\scriptsize
\setlength{\tabcolsep}{2pt}
\renewcommand{\arraystretch}{1.08}

\resizebox{\columnwidth}{!}{%
\begin{tabular}{@{}lccccc@{}}
\toprule
\textbf{Condition} &
\textbf{Perturped Base} &
\textbf{VLA-Corr.\cite{zju2026vlacorrector}} &
\textbf{A2C2 \cite{sendai2025a2c2}} &
\textbf{\spindle{}} &
$\boldsymbol{\Delta_{\mathrm{best}}}$ \\
\midrule
$\alpha=0.5$
& 37.0 (13.4) & 44.0 (15.6) & 52.0 (8.7)
& \textbf{63.7} (6.5) & $+11.7$ \\

$\alpha\sim\mathcal{U}[0.3,0.7]$
& 35.3 (14.1) & 43.7 (15.9) & 50.3 (9.1)
& \textbf{62.0} (6.7) & $+11.7$ \\

$\alpha\sim\mathcal{U}[0.75,0.85]$ 
& 40.7 (12.1) & 54.3 (11.8) & 52.3 (8.8)
& \textbf{65.7} (6.3) & $+11.3$ \\

All-joints Friction (5 steps)
& 35.7 (14.0) & 48.0 (13.9) & 49.7 (9.2)
& \textbf{64.7} (6.4) & $+15.0$ \\

All-joints Friction (sustained)
& 0.0 (--) & 0.3 (--) & 1.3 (--)
& \textbf{18.3} (30.3) & $+17.0$ \\
\bottomrule
\end{tabular}%
}

\vspace{1pt}
% \parbox{\columnwidth}{\scriptsize\raggedright
% Queries are omitted when fewer than five successes.

% }
\vspace{-2mm}
\end{table}

Having established \spindle{}'s robustness across actuator attenuation and joint friction, we next compare its recovery performance with two recent execution-time correctors for action-chunked VLAs: VLA-Corrector~\cite{zju2026vlacorrector} and A2C2~\cite{sendai2025a2c2}. As shown in Table~\ref{tab:correctors}, \spindle{} achieves the \textit{highest} success rate under every evaluated disturbance, outperforming the strongest competing corrector by 11.3--11.7,pp under actuator attenuation, 15.0,pp under transient friction, and 17.0,pp under sustained friction. The difference is particularly pronounced under sustained friction, where success falls to 0.3\% with VLA-Corrector and 1.3\% with A2C2, while \spindle{} retains 18.3\% success. This improved recovery is also achieved with fewer VLA queries: across the four conditions, \spindle{} requires only 6.3--6.7 queries per success, compared with 11.8--15.9 for VLA-Corrector and 8.7--9.2 for A2C2. Thus, \spindle{} provides stronger disturbance recovery while preserving the policy's original query schedule and requiring no additional corrective VLA queries.

\subsection{Ablation of the Learned Residual Predictor} \label{sec:results:residual}
We examine the contribution of the learned residual predictor $\rho_\theta$ by removing it from $F_\theta$ while retaining the calibrated analytical next-state estimate. Our results demonstrate that the learned residual is important for reliable execution correction under different disturbances. As shown in Table~\ref{tab:residual_ablation}, the analytical-only variant substantially degrades performance, with full \spindle{} outperforming it by 31.2~pp on average across the evaluated transient disturbances. The learned residual is also critical under clean execution as removing $\rho_\theta$ reduces success sharply from 66.7\% to 35.0\%, to even below the 64.0\% uncorrected baseline. These results indicate that the learned residual contributes to reliable execution-shortfall estimation beyond analytical compensation alone.

\begin{table}[t] \centering 
\caption{Ablation of the learned residual predictor on SmolVLA-T over 300 held-out contexts per condition. ``Analytical-only'' retains the calibrated analytical next-state estimate $\bar{x}_{t+1}$ but removes the learned residual predictor $\rho_\theta$; full \spindle{} combines both components. Friction here refers to all-joint friction. Entries report task success (\%).}
\label{tab:residual_ablation} \footnotesize \setlength{\tabcolsep}{12pt} \renewcommand{\arraystretch}{1.08} \begin{tabular}{lccc} \toprule Condition & Base & Analytical-only &  \spindle{} \\ \midrule No fault & 64.0 & 35.0 & \textbf{66.7} \\ $\alpha=0.5$ & 37.0 & 29.7 & \textbf{63.7} \\ $\alpha\sim\mathcal{U}[0.3,0.7]$ & 35.3 & 31.7 & \textbf{62.0} \\ $\alpha\sim\mathcal{U}[0.75,0.85]$ & 40.7 & 38.3 & \textbf{65.7} \\ Friction (5 steps) & 35.7 & 31.7 & \textbf{64.7} \\ Friction, sustained & 0.0 & 16.7 & \textbf{18.3} \\ \bottomrule \end{tabular} \end{table}

\subsection{Real-world Deployment and Experiments}
\label{sec:results:realworld}
% \textcolor{red}{Section DONE, don't touch}

While high-end industrial manipulators can come with significant cost overheads, 
ubiquitous real-world VLA deployment is increasingly targeting accessible, low-cost hardware, which naturally suffer from increased mechanical imperfections \cite{Kotov2026Backlash}. We therefore evaluate \spindle{} under natural hardware non-idealities on the real-world myCobot 320 object-placement setup following the real-world setup described in Section~\ref{sec:setup:realworld}.

% \begin{figure}[h]
% \centering
% \includegraphics[width=\columnwidth]{figs/illustration_robot.PNG}
% \caption{\textbf{Qualitative recovery example on the 6-DOF
% myCobot~320.} Without \spindle{} (top), execution drift causes a missed grasp,
% followed by a subsequent recovery attempt. With \spindle{} (bottom),
% within-chunk proprioceptive correction mitigates the execution error, enabling
% a successful grasp and task completion.}
% \label{fig:robotrealworldsnap}
% \vspace{-8pt}
% \end{figure}
\begin{table}[]
\centering
\caption{Real-world execution cost on the myCobot 320. }
\label{tab:hwcost}
\small
\setlength{\tabcolsep}{3pt}
\begin{tabular}{l l c c c}
\hline
Backbone & Method & Succ. (\%) & \shortstack{Queries /\\success} & \shortstack{Steps /\\success} \\
\hline
SmolVLA-T & Base      & 55.0 & 10.2 & 409 \\
           & +\spindle{} & 61.7 &  8.5 & 340 \\
SmolVLA-M & Base      & 35.0 & 20.2 & 811 \\
           & +\spindle{} & 48.3 & 13.1 & 530 \\
\hline
\end{tabular}
\end{table}

% While high-end industrial manipulators can come with significant cost overheads , 
% ubiquitous real-world VLA deployment is increasingly targeting accessible, low-cost hardware, which naturally suffer from increased mechanical imperfections \cite{Kotov2026Backlash}. 
% To validate our \spindle{} approach under real-world mechanical imperfections, we deploy our \textit{SmolVLA-T} and \textit{SmolVLA-M} models on a 6-DOF \textit{myCobot 320} arm executing an object placement 
% task. The control operates at a frame rate of $10~\text{Hz}$ with an execution horizon of $N=50$, leading to $\sim 5$ seconds of \textit{open-loop} execution per VLA query.

% \begin{figure}[h]
% \centering
% \includegraphics[width=\linewidth]{figs/robot_deployment_600dpi.pdf}
% \caption{Physical myCobot 320 tabletop pick-and-place
% performance across 120 evaluation runs per policy (60
% runs uncorrected vs. 60 runs +Spindle). }
% \label{fig:physres}
% \end{figure}

\subsubsection{Effect of \spindle{} on the alleviation of real-world non-idealities}

% \textcolor{orange}{Physical deployments show the same directional improvement observed in simulation, increasing placement success from 55.0\% to 61.7\% on SmolVLA-T and from 35.0\% to 48.3\% under zero-shot transfer to SmolVLA-M}
Here, evaluate whether \spindle{} mitigates the performance degradation arising from natural hardware non-idealities on the low-cost myCobot~320. Across 60 trials  per condition, the uncorrected baseline policies exhibit noticeable performance degradation due to inherent hardware non-idealities, achieving placement success rates of 55.0\% on SmolVLA-T and 35.0\% on SmolVLA-M (see Table. \ref{tab:hwcost} ), showing significant degradations compared to the base success rates attained \textit{in simulation} and reported in Table \ref{tab:baseresults}. Then, enabling \spindle{} raises placement success to 61.7\% (+6.7\,pp) for SmolVLA-T and 48.3\% (+13.3\,pp) zero-shot for SmolVLA-M (Table \ref{tab:hwcost} a). These results demonstrate that the proposed execution-level correction remains effective under natural real-world hardware non-idealities. Qualitatively, Fig.~\ref{fig:1} illustrates a representative object-picking sequence in which \spindle{} corrects execution drift that otherwise propagates to a failed grasp.

% Qualitatively, Fig. \ref{fig:robotrealworldsnap} provides snapshots of a real-world object picking sequence comparing the system behaviors with and without \spindle{}, clearly showing the correction ability of the proposed method in the illustrated case.

% When the models are used \textit{without} \spindle{}, (60 trials per configuration), uncorrected baseline policies exhibit noticeable performance degradation due to inherent joint gear backlash, motor latency, and friction, achieving placement success rates of 55.0\% on SmolVLA-T and 35.0\% on SmolVLA-M (see Table. \ref{tab:hwcost} ), showing significant degradations compared to the base success rates attained \textit{in simulation} and reported in Table \ref{tab:baseresults}. Then, adding \spindle{} consistently elevates the task success  to 61.7\% (+6.7\,pp) for SmolVLA-T and 48.3\% (+13.3\,pp) zero-shot for SmolVLA-M (Table \ref{tab:hwcost} a). These results further confirm the usefulness of the proposed approach for disturbance alleviation in the real-world context. Qualitatively, Fig. \ref{fig:robotrealworldsnap} provides snapshots of a real-world object picking sequence comparing the system behaviors with and without \spindle{}, clearly showing the correction ability of the proposed method in the illustrated case.

\subsubsection{Query Efficiency on the Arm}
\label{sec:results:hwcost}

The directional success gains are accompanied  by lower high-level inference cost per successful physical placement (Table~\ref{tab:hwcost}). Queries per successful placement decrease
from $10.2$ to $8.5$ for SmolVLA-T and from $20.2$ to $13.1$ for SmolVLA-M;
control steps per success decrease from $409$ to $340$ and from $811$ to $530$,
respectively. These reductions do not arise from modifying the scheduled VLA
query interval. Instead, successful corrected rollouts terminate with fewer
replanning cycles, while  our low-level predictor  adds approximately
$0.4$~ms of CPU time to each $100$~ms control step, or less than 1\% of its total time.

\subsection{Discussion}
\spindle{} improves the robustness of frozen, action-chunked VLAs at the execution layer: instead of querying the policy more often, it adds control-rate proprioceptive feedback within the committed chunk, without modifying the policy or issuing extra queries. Because it repays unrealized motion, its gains are largest at long open-loop horizons, where accumulated execution error matters most, so it helps preserve the inference-efficiency benefit of long chunks. This is most relevant to low-cost platforms, where backlash, tracking error, and mechanical variability can degrade otherwise valid VLA actions; on the myCobot~320, \spindle{} gave directionally higher placement success with fewer VLA queries per successful task.

\section{Conclusion}
\label{sec:conclusion}

We presented \spindle{}, a lightweight proprioceptive correction framework for robust execution of frozen, action-chunked VLAs. \spindle{} combines calibrated actuation modeling with learned nominal-execution prediction to detect execution shortfalls and correct subsequent actions within the same chunk, without VLA fine-tuning, chunk regeneration, or additional corrective VLA queries. Across three VLA backbones on LIBERO-Object, \spindle{} recovers much of the success lost to actuator attenuation and transient joint friction, gives a more favorable success--query trade-off at long execution horizons, and outperforms two recent execution-time correctors on SmolVLA-T. On a low-cost myCobot 320, it ran within the control budget and raised placement success directionally, with fewer VLA queries per successful task. These results support lightweight proprioceptive feedback as a practical means of improving VLA execution robustness without changing the policy or its query schedule. \spindle{} targets execution-level shortfalls rather than semantic policy errors or timing failures; extending it across task suites and embodiments, and testing it under injected faults on hardware, remain future work.

% In this letter, we presented \spindle{}, a lightweight proprioceptive correction framework for robust execution of frozen, action-chunked VLAs. \spindle{} combines calibrated actuation modeling with learned nominal-execution prediction to identify physical execution shortfalls and causally correct subsequent actions within the same chunk, without VLA fine-tuning, chunk regeneration, or additional corrective policy queries. Simulation experiments across three VLA backbones demonstrate substantial recovery under diverse execution disturbances, improved success--query trade-offs at longer execution horizons, and stronger recovery than the evaluated VLA correctors. Further, physical deployment on a low-cost myCobot 320 improves placement success and reduces the number of VLA queries required per successful task. Overall, these results show that lightweight proprioceptive feedback can improve the physical execution robustness of action-chunked VLAs while preserving their inference structure, particularly for deployment on mechanically imperfect, low-cost robotic platforms. The current formulation remains focused on execution-level shortfall rather than semantic policy errors or broader timing failures; extending this correction principle across additional robot embodiments and and more diverse real-world tasks are  future directions for work.

\bibliographystyle{IEEEtran}
\bibliography{refs_v3}

@article{sendai2025a2c2,
  author  = {Sendai, Kohei and Alvarez, Maxime and Matsushima, Tatsuya and
             Matsuo, Yutaka and Iwasawa, Yusuke},
  title   = {Leave No Observation Behind: Real-time Correction for {VLA}
             Action Chunks},
  journal = {arXiv preprint arXiv:2509.23224},
  year    = {2025}
}

@article{zju2026vlacorrector,
  author  = {Pan, Yi and Pan, Miao and Lu, Qi and Huang, Jiaming and
             Zhang, Man and Huang, Siteng and Li, Xin and Zhang, Jie and
             Shen, Yongliang and Zhang, Xuhong and Zhang, Wenqi},
  title   = {{VLA-Corrector}: Lightweight Detect-and-Correct Inference for
             Adaptive Action Horizon},
  journal = {arXiv preprint arXiv:2607.01804},
  year    = {2026}
}

@inproceedings{black2025rtc,
  author    = {Black, Kevin and Galliker, Manuel Y. and Levine, Sergey},
  title     = {Real-Time Execution of Action Chunking Flow Policies},
  booktitle = {Advances in Neural Information Processing Systems (NeurIPS)},
  year      = {2025}
}

@inproceedings{liu2025bid,
  author    = {Liu, Yuejiang and Hamid, Jubayer Ibn and Xie, Annie and Lee,
               Yoonho and Du, Maximilian and Finn, Chelsea},
  title     = {Bidirectional Decoding: Improving Action Chunking via Guided
               Test-Time Sampling},
  booktitle = {International Conference on Learning Representations (ICLR)},
  year      = {2025}
}

@inproceedings{ankile2025resip,
  author    = {Ankile, Lars and Simeonov, Anthony and Shenfeld, Idan and
               Torne, Marcel and Agrawal, Pulkit},
  title     = {From Imitation to Refinement -- Residual {RL} for Precise
               Assembly},
  booktitle = {IEEE International Conference on Robotics and Automation
               (ICRA)},
  year      = {2025}
}

@article{kim2026objectcentric,
  author  = {Kim, Kinam and Saito, Namiko and Kim, Heecheol and Ikeuchi,
             Katsushi and Choo, Jaegul and Matsushita, Yasuyuki},
  title   = {Object-Centric Residual {RL} for Zero-Shot Sim-to-Real {VLA}
             Enhancement},
  journal = {arXiv preprint arXiv:2606.18953},
  year    = {2026}
}

@article{jo2026jparc,
  author  = {Jo, Minsoo and Kwon, Taeju and Chun, Junha and Jeong, Youngjoon
             and Kim, Taesup},
  title   = {Uncovering Vulnerability of Vision-Language-Action Models under
             Joint-Level Physical Faults},
  journal = {arXiv preprint arXiv:2606.10501},
  year    = {2026}
}

@article{zhang2026ppc,
  author  = {Zhang, Yanyan and Song, Chaoda and Singh, Vikash and Li, Xinpeng
             and Ye, Kai and Hu, Zhe and Pu, Zhongzhu and Yin, Yu and
             Chaudhary, Vipin},
  title   = {Overcoming Dynamics-Blindness: Training-Free Pace-and-Path
             Correction for {VLA} Models},
  journal = {arXiv preprint arXiv:2605.11459},
  year    = {2026}
}

@article{physvla2026,
  author  = {Chandra, Namai and Damodaran, Shriram and Wang, Lin},
  title   = {{PhysVLA}: Towards Physically-Grounded {VLA} for Embodied
             Robotic Manipulation},
  journal = {arXiv preprint arXiv:2606.13886},
  year    = {2026}
}

@article{ohnishi1987,
  author  = {Ohnishi, Kouhei},
  title   = {A New Servo Method in Mechatronics},
  journal = {Transactions of the Japanese Society of Electrical Engineers},
  volume  = {107-D},
  pages   = {83--86},
  year    = {1987}
}

@article{chen2016dobc,
  author  = {Chen, Wen-Hua and Yang, Jun and Guo, Lei and Li, Shihua},
  title   = {Disturbance-Observer-Based Control and Related Methods: An
             Overview},
  journal = {IEEE Transactions on Industrial Electronics},
  volume  = {63},
  number  = {2},
  pages   = {1083--1095},
  year    = {2016}
}

@article{tang2020dob,
  author  = {Tang, Chen and Xu, Zhuo and Tomizuka, Masayoshi},
  title   = {Disturbance-Observer-Based Tracking Controller for Neural
             Network Driving Policy Transfer},
  journal = {IEEE Transactions on Intelligent Transportation Systems},
  volume  = {21},
  number  = {9},
  pages   = {3961--3972},
  year    = {2020}
}

@article{zhao2023act,
  author  = {Zhao, Tony Z. and Kumar, Vikash and Levine, Sergey and Finn,
             Chelsea},
  title   = {Learning Fine-Grained Bimanual Manipulation with Low-Cost
             Hardware},
  journal = {arXiv preprint arXiv:2304.13705},
  year    = {2023}
}

@inproceedings{chi2023diffusion,
  author    = {Chi, Cheng and Xu, Zhenjia and Feng, Siyuan and Cousineau,
               Eric and Du, Yilun and Burchfiel, Benjamin and Tedrake, Russ
               and Song, Shuran},
  title     = {Diffusion Policy: Visuomotor Policy Learning via Action
               Diffusion},
  booktitle = {Robotics: Science and Systems (RSS)},
  year      = {2023}
}

@article{shukor2025smolvla,
  author  = {Shukor, Mustafa and Aubakirova, Dana and Capuano, Francesco and
             Kooijmans, Pepijn and Palma, Steven and Zouitine, Adil and
             Aractingi, Michel and Pascal, Caroline and Russi, Martino and
             Marafioti, Andres and Alibert, Simon and Cord, Matthieu and
             Wolf, Thomas and Cadene, Remi},
  title   = {{SmolVLA}: A Vision-Language-Action Model for Affordable and
             Efficient Robotics},
  journal = {arXiv preprint arXiv:2506.01844},
  year    = {2025}
}

@inproceedings{liu2023libero,
  author    = {Liu, Bo and Zhu, Yifeng and Gao, Chongkai and Feng, Yihao and
               Liu, Qiang and Zhu, Yuke and Stone, Peter},
  title     = {{LIBERO}: Benchmarking Knowledge Transfer for Lifelong Robot
               Learning},
  booktitle = {Advances in Neural Information Processing Systems (NeurIPS),
               Datasets and Benchmarks Track},
  year      = {2023}
}

@article{intelligence2025pi05,
  author  = {{Physical Intelligence} and Black, Kevin and Brown, Noah and
             Darpinian, James and Dhabalia, Karan and Driess, Danny and
             others},
  title   = {$\pi_{0.5}$: A Vision-Language-Action Model with Open-World
             Generalization},
  journal = {arXiv preprint arXiv:2504.16054},
  year    = {2025}
}

@article{black2024pi0,
  author  = {Black, Kevin and Brown, Noah and Driess, Danny and Esmail,
             Adnan and Equi, Michael and Finn, Chelsea and others},
  title   = {$\pi_0$: A Vision-Language-Action Flow Model for General Robot
             Control},
  journal = {arXiv preprint arXiv:2410.24164},
  year    = {2024}
}

@article{kim2025oft,
  author  = {Kim, Moo Jin and Finn, Chelsea and Liang, Percy},
  title   = {Fine-Tuning Vision-Language-Action Models: Optimizing Speed and
             Success},
  journal = {arXiv preprint arXiv:2502.19645},
  year    = {2025}
}

@article{mohsen2026mamba,
  title={Mamba-based Selective State Space Modeling Improves the Accuracy-Complexity Tradeoff of SmolVLA Vision-Language-Action Experts},
  author={Mohsen, Farida and Elkaffash, Thowayba and Qazani, Mohammad Reza Chalak and Mabrok, Mohamed and Meskin, Nader and Safa, Ali},
  journal={arXiv preprint arXiv:2608.21407},
  year={2026}
}

@article{zhu2020robosuite,
  author  = {Zhu, Yuke and Wong, Josiah and Mandlekar, Ajay and
             Mart{\'i}n-Mart{\'i}n, Roberto and Joshi, Abhishek and Lin,
             Kevin and Maddukuri, Abhiram and Nasiriany, Soroush and Zhu,
             Yifeng},
  title   = {robosuite: A Modular Simulation Framework and Benchmark for
             Robot Learning},
  journal = {arXiv preprint arXiv:2009.12293},
  year    = {2020}
}

@inproceedings{zhou2019rotation,
  author    = {Zhou, Yi and Barnes, Connelly and Lu, Jingwan and Yang, Jimei
               and Li, Hao},
  title     = {On the Continuity of Rotation Representations in Neural
               Networks},
  booktitle = {IEEE/CVF Conference on Computer Vision and Pattern Recognition
               (CVPR)},
  pages     = {5738--5746},
  year      = {2019}
}

@article{zhao2025vlarail,
  title={VLA-RAIL: A Real-Time Asynchronous Inference Linker for VLA Models and Robots},
  author={Zhao, Yongsheng and Zhao, Lei and Cheng, Baoping and Yao, Gongxin and Wen, Xuanzhang and Gao, Han},
  journal={arXiv preprint arXiv:2512.24673},
  year={2025}
}

@inproceedings{song2025pdvla,
  title={PD-VLA: Accelerating Vision-Language-Action Model Integrated with Action Chunking via Parallel Decoding},
  author={Song, Wenxuan and Chen, Jiayi and Ding, Pengxiang and Zhao, Han and Zhao, Wei and Zhong, Zhide and Ge, Zongyuan and Li, Zhijun and Wang, Donglin and Wang, Lujia and Ma, Jun and Li, Haoang},
  booktitle={2025 IEEE/RSJ International Conference on Intelligent Robots and Systems (IROS)},
  pages={13162--13169},
  year={2025},
  publisher={IEEE},
  doi={10.1109/IROS60139.2025.11247519}
}

@article{din2025vision,
  title={Vision language action models in robotic manipulation: A systematic review},
  author={Din, Muhayy Ud and Akram, Waseem and Saoud, Lyes Saad and Rosell, Jan and Hussain, Irfan},
  journal={arXiv preprint arXiv:2507.10672},
  year={2025}
}

@inproceedings{chopra2026everydayvla,
  title={Everydayvla: A vision-language-action model for affordable robotic manipulation},
  author={Chopra, Samarth and McMoil, Alex and Carnovale, Ben and Sokolson, Evan and Kubendran, Rajkumar and Dickerson, Samuel},
  booktitle={2026 IEEE International Conference on Robotics and Automation (ICRA)},
  pages={677--684},
  year={2026},
  organization={IEEE}
}

@inproceedings{deisenroth2011lowcost,
  author    = {Marc Peter Deisenroth and Carl Edward Rasmussen and Dieter Fox},
  title     = {Learning to Control a Low-Cost Manipulator using Data-Efficient Reinforcement Learning},
  booktitle = {Proceedings of Robotics: Science and Systems (RSS)},
  address   = {Los Angeles, CA, USA},
  year      = {2011},
  doi       = {10.15607/RSS.2011.VII.008}
}

@article{murali2019pyrobot,
  author  = {Adithyavairavan Murali and Tao Chen and Kalyan Vasudev Alwala and Dhiraj Gandhi and Lerrel Pinto and Saurabh Gupta and Abhinav Gupta},
  title   = {{PyRobot}: An Open-source Robotics Framework for Research and Benchmarking},
  journal = {arXiv preprint arXiv:1906.08236},
  year    = {2019}
}

@article{yang2026fpc,
  title={Fpc-vla: A vision-language-action framework with a supervisor for failure prediction and correction},
  author={Yang, Yifan and Duan, Zhixiang and Xie, Tianshi and Cao, Fuyu and Shen, Pinxi and Song, Peili and Zhao, Chenyang and Jin, Piaopiao and Sun, Guokang and Xu, Shaoqing and others},
  journal={Expert Systems with Applications},
  pages={131742},
  year={2026},
  publisher={Elsevier}
}

@article{song2026geobridge,
  title={GeoBridge-VLA: Geometry-Aware Residual Adaptation for Vision-Language-Action Models},
  author={Song, Hyun and Kim, Kangmin and Um, Loren Jinsoo and Han, Minhui and Park, Jaehyeok and Cho, Taewan and Choi, Andrew Jaeyong},
  journal={arXiv preprint arXiv:2610.05026},
  year={2026}
}

@inproceedings{dalal2025local,
  title={Local policies enable zero-shot long-horizon manipulation},
  author={Dalal, Murtaza and Liu, Min and Talbott, Walter and Chen, Chen and Pathak, Deepak and Zhang, Jian and Salakhutdinov, Ruslan},
  booktitle={2025 IEEE International Conference on Robotics and Automation (ICRA)},
  pages={13875--13882},
  year={2025},
  organization={IEEE}
}

@article{lu2026would,
  title={When would vision-proprioception policies fail in robotic manipulation?},
  author={Lu, Jingxian and Xia, Wenke and Wu, Yuxuan and Lu, Zhiwu and Hu, Di},
  journal={arXiv preprint arXiv:2602.12032},
  year={2026}
}

@article{li2026revip,
  title={ReViP: Mitigating False Completion in Vision-Language-Action Models with Vision-Proprioception Rebalance},
  author={Li, Zhuohao and Li, Yinghao and Jiang, Jian-Jian and Zhou, Lang and Zhang, Tianyu and Yin, Jiadong and Lin, Mu and Wei, Yi-Lin and Zheng, Wei-Shi},
  journal={arXiv preprint arXiv:2601.16667},
  year={2026}
}

@article{liu2026palm,
  title={Palm: Progress-aware policy learning via affordance reasoning for long-horizon robotic manipulation},
  author={Liu, Yuanzhe and Zhu, Jingyuan and Mo, Yuchen and Li, Gen and Cao, Xu and Jin, Jin and Shen, Yifan and Li, Zhengyuan and Yu, Tianjiao and Yuan, Wenzhen and others},
  journal={arXiv preprint arXiv:2601.07060},
  year={2026}
}

@inproceedings{todorov2012mujoco,
  author    = {Todorov, Emanuel and Erez, Tom and Tassa, Yuval},
  title     = {{MuJoCo}: A physics engine for model-based control},
  booktitle = {Proc. IEEE/RSJ Int. Conf. Intelligent Robots and Systems (IROS)},
  pages     = {5026--5033},
  year      = {2012},
  doi       = {10.1109/IROS.2012.6386109}
}

@article{armstrong1994survey,
  author  = {Armstrong-H{\'e}louvry, Brian and Dupont, Pierre and
             {Canudas de Wit}, Carlos},
  title   = {A survey of models, analysis tools and compensation methods for
             the control of machines with friction},
  journal = {Automatica},
  volume  = {30},
  number  = {7},
  pages   = {1083--1138},
  year    = {1994},
  doi     = {10.1016/0005-1098(94)90209-7}
}

@article{kotov2026Backlash,
  author  = {Boris Kotov},
  title   = {An Open-Source Test Stand for Backlash Measurement in Low-Cost UART Servo Motors},
  journal = {HardwareX},
  volume  = {26},
  pages   = {e00767},
  year    = {2026},
  doi     = {10.1016/j.ohx.2026.e00767}
}

@article{dou2026neuralactuator,
  title={NeuralActuator: Neural Actuation Modeling for Robot Dynamics and External Force Perception},
  author={Dou, Zhiyang and Onyemelukwe, John U and Zhang, Hangxing and Zhang, Heng and Guo, Minghao and Tian, Yunsheng and Lipiec, Michal Piotr and Jacob, Joshua and Liu, Chao and Chen, Peter Yichen and others},
  journal={arXiv preprint arXiv:2607.11734},
  year={2026}
}

@book{barfoot2017state,
  author    = {Timothy D. Barfoot},
  title     = {State Estimation for Robotics},
  publisher = {Cambridge University Press},
  year      = {2017},
  doi       = {10.1017/9781316671528}
}

\end{document}